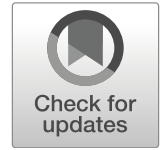

# LabelSens: enabling real-time sensor data labelling at the point of collection using an artificial intelligence-based approach

Kieran Woodward[1] · Eiman Kanjo[1] · Andreas Oikonomou[1] · Alan Chamberlain[2]

## Abstract
In recent years, machine learning has developed rapidly, enabling the development of applications with high levels of recognition accuracy relating to the use of speech and images. However, other types of data to which these models can be applied have not yet been explored as thoroughly. Labelling is an indispensable stage of data pre-processing that can be particularly challenging, especially when applied to single or multi-model real-time sensor data collection approaches. Currently, real-time sensor data labelling is an unwieldy process, with a limited range of tools available and poor performance characteristics, which can lead to the performance of the machine learning models being compromised. In this paper, we introduce new techniques for labelling at the point of collection coupled with a pilot study and a systematic performance comparison of two popular types of deep neural networks running on five custom built devices and a comparative mobile app (68.5–89% accuracy within-device GRU model, 92.8% highest LSTM model accuracy). These devices are designed to enable real-time labelling with various buttons, slide potentiometer and force sensors. This exploratory work illustrates several key features that inform the design of data collection tools that can help researchers select and apply appropriate labelling techniques to their work. We also identify common bottlenecks in each architecture and provide field tested guidelines to assist in building adaptive, high-performance edge solutions.

**Keywords** Labelling methods · Data · Machine learning · Artificial intelligence · AI · Multi-modal recognition · Pervasive computing · Tangible computing · Internet of things · IoT · HCI

## 1 Introduction

Deep neural networks (DNNs) are attracting more and more attention and are commonly seen as a breakthrough in the advance of artificial intelligence demonstrating DNNs' potential to be used to accurately classify sensory data. However, in order to train DNNs, vast quantities of data must first be collected and labelled. This data can include videos, images, audio, physical activity-related data, temperature and air quality, inevitably resulting in huge datasets containing data relating to all types of actions and behaviours. Labelling such data is not a trivial task, especially when the premise of such systems is to enable real-time machine learning, such as recognising emotions or security threats. So far, most of the attention has been focused on the processing power of edge computing devices [1, 2], and little attention has been paid on how to obtain clean and efficient labelled data to train models [3].

When collecting data in "the wild", in the real-world outside the confines of the research lab [4], a participant could be doing anything from driving a car to eating in a restaurant. Labelling can be either automatic or manual, which can be particularly challenging when people are engaged in physical activities. Taking this into account, the nature of each activity needs to be considered, both at a UX and user interface design level, as well as for data sources and providers, and at the application level.

It is crucial to label sensor data in real time, because unlike images and audio, it is not usually possible to label the data offline using the raw data itself without the real-time context. In pervasive sensing, there are three data collection methods [4]; these are (1) passive data sensing using smartphones or other sensors to record unlabelled data in the background [4]

✉ Alan Chamberlain
  alan.chamberlain@nottingham.ac.uk

[1] Nottingham Trent University, Nottingham, UK

[2] University of Nottingham, Nottingham, UK

Springer

often used to collect weather-related data [5], health data [6, 7] and environmental data [8]; (2) alternatively, active data sensing enables users to label the data in real time through self-reporting often; this is often used to report well-being or physical activity; (3) hybrid data sensing combines both passive and active data collection as it involves users actively labelling the passive sensor data that is recorded in the background [9].

The choice of a labelling approach depends on the complexity of a problem, the required training data, the size of a data science team and the financial and time resources a company can allocate to implement a project. The best approach to label data often fundamentally depends on the data and source type being recorded, e.g. sensor data can utilise mobile phone applications to collect labelled data whereas labelling images and audio may utilise post-processing techniques to implicitly crowdsource the labels such as Google's reCAPTCHA [10].

The labelling rate of sensor data can also dictate which approach to choose as data that frequently changes may require a higher labelling rate along with a more convenient labelling approach. The sample size is another factor that can dictate a labelling approach; the labelling of images can be automated or crowdsourced whereas a large sample size of sensor data requires recruiting many participants, for what could be an extended period of time. Crowdsourcing labels using web-based applications is often employed for images and audio data tagging, as it is most commonly processed offline [11]. This is not possible with time-series data which has to be labelled online in real time at the point of collection due to the nature of the data. Outsourcing the labelling of image, video and audio data to private companies is also gaining popularity although this is also not possible for sensor data as activities cannot be deduced from the raw data, meaning real-time labelling techniques must be developed [12].

Tangible user interfaces (TUIs) [13] present significant opportunities for the real-time labelling of sensor data. Tangible interfaces are physical interfaces that enable users to interact with digital information. These interfaces can be embedded with a variety of sensors including those which are not commonplace in smartphones such as physiological sensors (electrodermal activity (EDA), heart rate variability (HRV)) and environmental sensors (barometric pressure, ultraviolet (UV)) enabling the collection of in situ data for all sensors. TUIs can vary in size and shape but contain ample space to include the necessary sensors in addition to a real-time labelling technique.

To address the above challenges, we introduce LabelSens, a new framework for labelling sensor data at the point of collection. Our approach helps developers in adopting labelling techniques that can achieve higher performance levels. In this paper, we present five prototypes utilising different tangible labelling mechanisms and provide a comprehensive performance comparison and analysis of these prototypes. In particular, two popular deep learning networks were tested: long short-term memory (LSTM) and gated recurrent unit (GRU). Both were used to classify human generated, physiological activity data collected from 10 users.

Activity recognition is an established field; however, the methods used to label the sensor data collected are greatly underexplored. Researchers often manually label the activity participants undertake [14] which typically prevents the collection of in situ data as it requires the researcher to continuously video participants' activities so that the data might be labelled offline. Previous research has utilised smartphone applications to enable users to self-label their current activity using onscreen buttons [15]. However, it is not possible to use smartphones to collect data when additional sensors that are not embedded within smartphones are required, e.g. EDA or UV. It is possible for a combination of a smartphone application (for labelling), and YUIs (for sensory data collection) could be used, but this increases the complexity of the system by forcing users to use 2 devices; it also requires a continuous stable wireless connection between the 2 devices.

Little research has been conducted to evaluate the feasibility and performance of other real-time labelling techniques that would be suitable for edge devices. Looking beyond the data collection stage, we also start to examine the classification accuracy of different labelling techniques.

In this paper our contribution is two-fold; firstly, we introduce a novel mechanism to label sensory data on edge computing and TUIs while conducting a pilot study to collect training data for machine learning algorithms, and secondly we present a systematic way to assess the performance of these labelling mechanisms. Our results show that using LabelSens can be an effective method for collecting labelled data. The remainder of the paper is organized as follows: "Section 2" presents related work, while "Section 3" introduces our experimental methods. Labelling rate results are presented in "Section 4". In "Section 5", we present the algorithms used in the research; this is followed by the Discussion—"Section 6". Potential applications and future work are further explored and discussed in "Section 7", while "Section 8" concludes the paper.

## 2 Background

### 2.1 Data Labelling

As we have already seen, there are numerous approaches to labelling which vary depending on the data being collected. Sensor data is most commonly labelled using a hybrid approach where the sensor data is recorded continuously, while the user occasionally records a label against all or part of the previously recorded data. The labelling of human activities increasingly relies on hybrid data collection techniques using smartphones to continuously record accelerometer data as well as enabling users to self-report their current activity [15]. Smartphone applications are becoming increasingly popular to label sensor data as they provide a familiar, always accessible interface for users, although recently the use of new smartphone labelling techniques such as NFC and the use of volume buttons have proved to be an intuitive and popular approach when using an alternative approach is not possible [4]. Active learning [16] can be used to label data, needing few labelled training instances as the machine learning algorithm chooses the data from which it learns. Active learning could be beneficial for data where it is challenging to crowdsource labels, such as raw sensor data that is not sufficiently labelled [17]. Legion:AR [12] used the power of crowdsourcing combined with active learning to label human activities. Active learning was used to automate the labelling process where it was paired with real-time human labellers to label the data that could not be correctly labelled automatically. However, this approach requires cameras to continually record users in order that the unlabelled activities can be tagged offline. This may be feasible in specific scenarios, such as the workplace, but would not be plausible "in the wild". Another method to crowdsource human activities requires users to record short video clips of themselves performing different actions at home [18]. While crowdsourcing videos can result in ample data, it only allows for video data to be captured with no other sensor feeds and relies on the willingness of people to perform different activities on video.

As we have noted, the techniques used to label data vary depending on the data type as images can be labelled offline using an automated process based on click-through data, greatly reducing the effort required to create a labelled dataset [19]. Additionally, online tools have been developed that enable users to highlight and label objects within images. The use of an online tool allowed people from around the world to help label objects within images which are simply not possible with the sensor data [20].

Labelling audio data uses an approach similar to that of images, as spoken words are often labelled "in-house" by linguistic experts or may be crowdsourced. There are many forms of audio labelling including genre classification, vocal transcription and labelling various sounds within the audio, e.g. labelling where bird calls begin and finish. One labelling solution primarily focused on the artwork albums (recorded music), text reviews and audio tracks to label over 30,000 albums in relation to one of 250 labels provided, using deep learning to provide a related multi-label genre classification [21]. While labelling sounds can be crowdsourced, encouraging individuals to correctly label data can be a challenging task as it can be tedious. To increase compliance and engagement during labelling, previous research has developed games such as Moodswings [22] and TagATune [23] where players label different sounds. TagATune demonstrates the ability to engage users in labelling data as 10 out of 11 players said they were likely to play the game again.

Textual data from social media websites can be automatically labelled using the hashtags and emojis contained within posts as these often describe the contents of the post; however, this can result in noisy data [24]. Alternatively, text can be manually labelled but this is a labour-intensive process. One solution to this problem has involved training a machine learning model using a manually labelled dataset and then combining this with noisy emoticon data to refine the model through smoothing [25]. This method of combining labelled and noisy data outperformed models trained using just one data type.

## 2.2 Human machine interaction

The real-time labelling of sensor data is a more challenging proposition and often relies on the physical interaction with tangible interfaces. Recent advances in pervasive technologies have allowed engineers to transform bulky and inconvenient monitors into relatively small, comfortable and ergonomic research tools. Emoball [26] has been designed to enable users to self-label their mood by squeezing an electronic ball. While this device only allows users to report a limited number of emotions, participants said it was simple to use and liked the novel interaction approach. An alternative method to label mood was explored using a cube containing a face representing a different emotion of each face of the cube [27]. Users simply moved the cube to display the face that most represented their mood providing a simple, intuitive way for people to label data albeit limited by the number of faces on the cube. Mood TUI [28] goes beyond self-reporting to a hybrid approach in order for users to record their emotions and collect relevant data from the user's smartphone including location and physiological data such as heart rate. Participants found the use of TUIs very exciting, demonstrating the potential to increase the usability and engagement of labelling, but thus far, they have not been widely utilised outside of self-reporting emotions.

Numerous methods that used to self-report emotions have been explored including touch, motion and buttons. These interaction techniques have paved the way for unique interactions with devices, but limited sensor data has been recorded, and the accuracy of the techniques has not been evaluated as previous research has not used data collected for machine learning but purely as a method for individuals to self-report their well-being.

Sometimes it is not physically possible to interact with physical devices to label sensor data, such as when an individual is driving. A solution to this problem has been the use of the participants' voice, for example, to label potholes in the road [29]. When labelling rapidly changing data, such as road conditions, it can be difficult to label the data at the exact time when a label needs to be added, so techniques may be used to analyse sensor data windows near the label to allow the exact pothole readings to be correctly labelled. Techniques such as these are vital to ensure that the sensor data is correctly labelled as incorrectly labelled data will result in inaccurate machine learning models that will not be able to correctly classify any future data. However, vocal labelling is not practical if the device is to be used in public for task that use sensitive label data such as relating to emotional well-being or, for example, if there is a considerable amount of background noise in the environments that would interfere with the recognition levels.

Table 1 (*above*) shows the current labelling approaches used including in-house labelling and crowdsourced labelling, requiring user activities to be video recorded enabling offline manual data labelling. Similarly, automatic labelling can use large amounts of labelled video or sensor data to enable future data to be automatically labelled, dramatically reducing the time required to label but also reducing the accuracy in which the data is labelled. Alternatively, Generative Adversarial Networks (GAN) can be used to automatically generate further labelled data, but a vast labelled dataset is initially required, and the synthetic data labels may be highly inaccurate.

In comparison, labelling at the point of collection is highly accurate, because it is done real time, it is cost-effective, it is time-effective and it enables in situ data to be collected. Thus far, however, labelling at the point of collection has had limited use, the main area of use has been smartphone applications. There are numerous scenarios where labelling sensor data at the point of collection would result in the most effective and accurate data, but there is currently no established framework to accomplish this. When providing participants with tangible interfaces to collect a wide array of sensory data, embedding a labelling method directly into the device simplifies the labelling process and allows for numerous sensors that are not embedded within smartphones to be utilised. This concept creates a simple, tangible, easy to use method to label sensor data in real time and in situ, aiming to improve the quantity and reliability of labelled data and therefore increasing the accuracy of machine learning models which might be applied.

Overall, there are numerous possibilities for text, audio and images to be labelled offline, unlike raw sensor data which, as we have previously noted, must be labelled in real time. TUIs have previously been used to self-report, but the data is often not collected to train machine learning models, which has meant the accuracy and validity of the labelling techniques has never been evaluated. Human activity recognition has been well-researched, but the techniques to label the data have always either involved offline labelling or a mobile phone application which limits the availability of sensors. The use of tangible interfaces containing different labelling methods in addition to a wide range of sensors has not been considered but could aid the real-time collection of labelled data. This research aims to explore the impact that different labelling techniques embedded within TUIs have on the accuracy of labelling, label rate, usability and classification performance.

## 3 LabelSens framework

### 3.1 Configuration and system architecture

Labelling at the point of data collection provides many benefits, which include lower associated costs, reduced time (on-task) and the ability to label data in situ. TUIs present many opportunities to embed unique physical labelling techniques that may be easier to use than comparative virtual labelling techniques used to collect in situ labelled data. In addition, TUIs provide ideal interfaces to directly embed a magnitude of sensors, negating the need for participants to carry the sensors in as well as a separate labelling mechanism.

Table 1 Comparison of frequently used labelling techniques

| Labelling technique | | Data collection | Related work | Description | Accuracy | Time | Cost |
|---|---|---|---|---|---|---|---|
| Human | In-house labelling | Video | Activity recognition [14] | Labelling carried out by in house trained team | High | Long | Low |
| | Crown source labelling | Video | reCAPTCHA [10] | Labelling carried out by external third parties (not trained) | Low | Long | High |
| | Labelling at the point of collection | Mobile | Mobile app [30] [31] | Labelling carried out by the user in situ and in real time | High | Short | Low |
| Automatic | | Sensor/video | Fujitsu [32] | Generating time-series data automatically from a previous extended data collection period | Low | Short | Low |
| Synthetic data | | Sensor/video | GAN [33] | Generating synthetic labelled dataset with similar attributes recently using Generative Adversarial Networks | Very low | Short | Low |

TUIs can vary in shape and size ranging from small wearables to physical devices that are designed to be frequently interacted with such as stress balls embedding force sensitive resistors to measure touch. This enables a wide array of opportunities to embed sensors within a variety of objects and combined with machine learning classifiers that could be used to infer behaviour change, emotions, movement and more. However, before machine learning models can be trained, a vast amount of labelled data is first required. By embedding a labelling technique along with the sensors within TUIs, it ensures the sensor data and label are both being collected in real time aiming to improve data collection rates and accuracy.

Figure 1 demonstrates the concept of the LabelSens framework, pairing time-series sensor data with a physical labelling technique inside a TUI to collect in situ labelled sensor data.

### 3.2 Labelling mechanisms

To understand the feasibility of labelling techniques for TUIs, we propose a range of alternative approaches to traditional labelling techniques. In this section, we present five new prototypes that each contain a unique labelling technique and will be used to label human activity (walking, climbing downstairs and climbing upstairs) along with a comparative mobile application:

- Two adjacent buttons (press one button for climbing upstairs, press the other button for climbing downstairs and press both buttons simultaneously to record walking)
- Two opposite buttons (press one button for climbing upstairs, press the other button for climbing downstairs and press both buttons simultaneously to record walking)
- Three buttons (one button each for climbing upstairs, climbing downstairs and walking)
- Force sensitive resistor to measure touch (light touch for walking, medium touch for climbing downstairs, hard touch for climbing upstairs)
- Slide potentiometer (slide to the left for climbing downstairs, slide to the middle for walking and slide to the right for climbing upstairs)
- An Android mobile application provided on a Google Pixel 3 smartphone with 3 virtual buttons to label walking, climbing downstairs and climbing upstairs

Each TUI is a 6 cm × 6 cm × 6 cm 3d printed cube that contains a labelling technique combined with the required sensor and microcontroller. The size of the TUI could be reduced dependent on the labelling technique used and the sensors required, but here all interfaces were on the same size to reduce bias. The embedded electronics include:

- Arduino Nano microcontroller. Due to its small size, being open source and being compatible with a range of sensors.
- Inertial measurement unit (IMU). To record motion data. An IMU with 9 degrees of freedom has been used as it integrates sensors: an accelerometer, a magnetometer and a gyroscope to provide better accuracy, adding additional data.
- Micro SD card reader to locally record the IMU sensor data along with the user inputted label.

The buttons and slide potentiometer enable users to easily visualise the activity they are labelling; when using the touch sensor, it is difficult to distinguish between the three levels of force. To visualise the selected label, a multicoloured LED has also been incorporated into the device that changes from green to yellow to red when the device is touched with low, medium and high force. Figure 2 shows the electronic circuit and the developed TUI for the three buttons labelling and slider interfaces.

The mobile application was developed for the Android operating system and was tested using a Google Pixel 3. The application consisted of three virtual buttons in the centre of the screen labelled *downstairs*, *walking* and *upstairs* when a button is pressed the text at the top of the screen changes to show the currently selected label. Finally, at the bottom of the screen are the two additional virtual buttons to begin and end the recording of data. The sensor data along with its label is then saved to a CSV file stored on the phone's internal storage. A challenge when the mobile app for data labelling was developed was the frequency of the data, as the gyroscopic data had a significantly lower frequency than the accelerometer data resulting in the reduction of data sampling frequency.

We envision TUIs being used to label a maximum of 5 classes to ensure users are not overwhelmed and can sufficiently label at all times. Additional buttons could be added, e.g. 1 button each for up to 5 classes, but as we are only classifying 3 activities, the impact of having varying number of buttons (2 or 3) can be explored. This novel approach to in situ labelling provides an easy to use interface that facilitates the collection of real-time labelled data. The mobile app presents easier opportunities to include more labels, but users may still be overwhelmed by numerous virtual buttons. Overall, the five prototypes demonstrate the variety of labelling techniques that can be used in comparison to traditional app based or offline labelling.

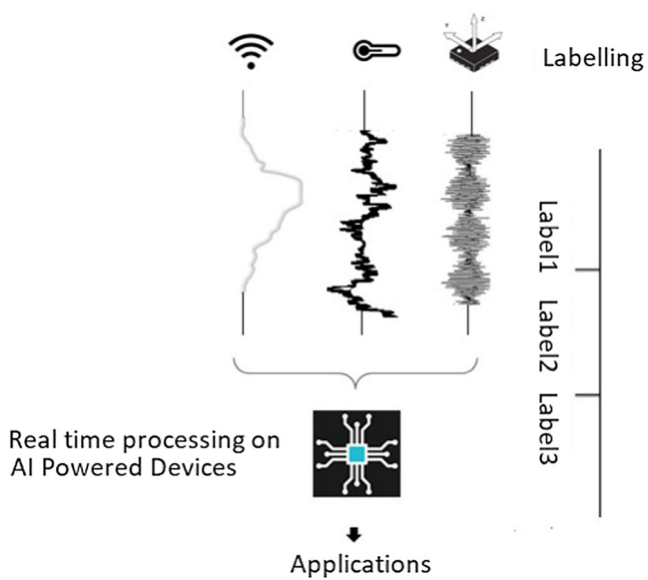

**Fig. 1** LabelSens framework: real-time sensor data fused with a label

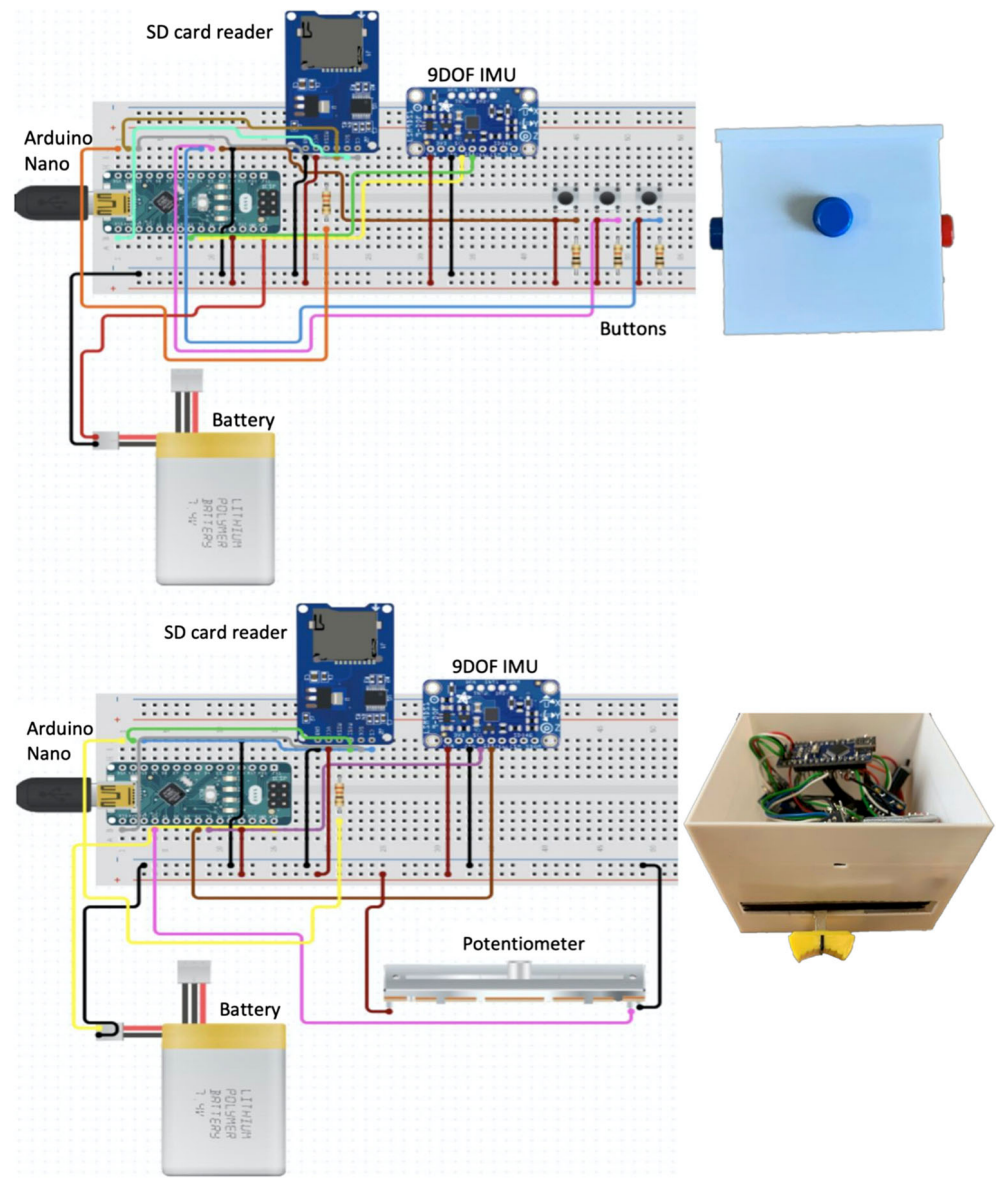

**Fig. 2** Example of two electronic circuits and developed tangible interface with three buttons and slider labelling interfaces

### 3.3 Experimental setup

An experiment was designed and conducted that explored the feasibility of the different self-labelling techniques in the interfaces. This pilot study involved ten participants who were initially shown a set route to follow to ensure sufficient data was collected for all three activities. Participants were instructed that the label should only be recorded when commencing a new activity, and if an incorrect label is recorded, then the correct label should be recorded as soon as possible to simulate real-world labelling. Each participant then used all of the interfaces containing the 5 different labelling techniques and the mobile apps for 3 min each while undertaking 3 activities: *walking, climbing upstairs* and *climbing downstairs*. Ideally the labelling system should be unobtrusive, in a way that the process of labelling the data should not alter of that affect the data being collected. Therefore, participants were not accompanied during the data collection period to realistically simulate in situ data collection which is the fundamental purpose of these interfaces. No issues arose during the data collection with each user understanding how to use each of the interfaces and successfully collecting data from all devices. The three activities were allowed for each participant to experience the different labelling techniques as well as collect sensor data which can be used to examine the accuracy and performance of each labelling technique.

## 4 Examining labelling rate

The maximum labelling rate of the devices is a key factor in deciding a labelling technique as some forms of sensor data can frequently change requiring a new label to be recorded multiple times every minute. To measure the maximum rate at which it is possible to label data, each interface was used continuously for 2 min to record the maximum number of label changes as possible. Figure 3 shows the total number of times each label was recorded on each of the devices.

The devices with only 2 buttons show the lowest data rate for each of the three labels because of a delay that was required to prevent mislabelling when simultaneously clicking both buttons to record the third label. The delay ensures that if a user releases one button slightly before the other when pressing both buttons to record the third label, the third label will still be recorded rather than the label for the button released last. The app shows a higher labelling rate than the devices with two buttons but is not significantly greater due to the difficulty in pressing virtual buttons that can easily be missed compared with physical buttons.

Three buttons show significantly more data recorded although very little data was recorded for one of the buttons possibly due to the third button being more difficult to reach as each button is located on a different face of the cube. The touch sensor recorded a high labelling rate for all three labels as to reach label 2 (high setting); by pressing the sensor, the user must first record label 0 and 1 as they increase the force exhorted on the sensor. The slider shows high labelling rates for label 0 and label 2 but not label 1 because it is simple to slide the slider from one end to the other, but the slider was rarely located in the middle of the device long enough for the label to be recorded. This shows the touch and slider techniques are easy to label the extreme values, but intermediary values are more challenging to frequently label. If all labels need to be frequently labelled, then buttons may be the best labelling technique although the position of the buttons can greatly impact the ease of which labelling can occur.

It is also vital to compare the number of times the label changed over the 2-min period to evaluate how simple it is to change label for each technique. Figure 4 shows the slider recorded the most label changes overall because of the simplicity to navigate between the labels followed by two opposite buttons which is surprising due to its low labelling rate. This demonstrates that while the use of buttons does not result in the highest labelling rate, it is simple to switch between the different labels and should be used when the label will change frequently. Touch, three buttons, the mobile app, and two adjacent buttons all performed similarly well showing there is little difference in accessing all of the labels when using these devices.

Once all the participants used each device to label while *walking, climbing downstairs and climbing upstairs*, the data was extracted, enabling comparisons to be established as shown in Fig. 4. The rate at which labels were changed from one label to another during the collection of activity data shows that the three buttons recorded the fewest in situ labelling changes for all users, while the two opposite buttons had the highest overall rate of in situ labelling changes albeit much lower than the maximum rate of labelling changes demonstrating fewer buttons increased ease of use. Labelling via touch had a consistently high rate of labelling changes for users, but this again could be due to the requirement of looping through all of the labels to reach the desired label. The mobile app achieved a slightly higher rate than three buttons and slider but not as high as the two buttons or touch. Overall the slider and the three buttons produced the lowest rate of label changes during data collection showing these labelling techniques should not be utilised with data that requires frequent labelling changes because of the difficulty in accessing all the three labels.

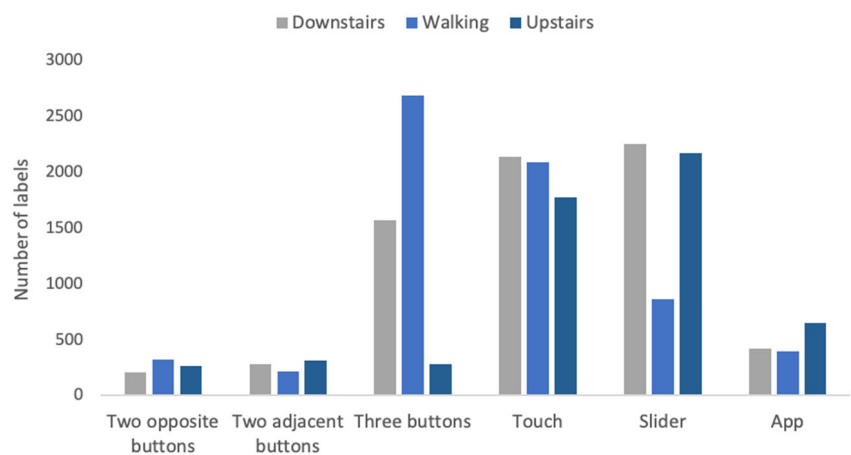

**Fig. 3** Maximum labelling rate for each label per device

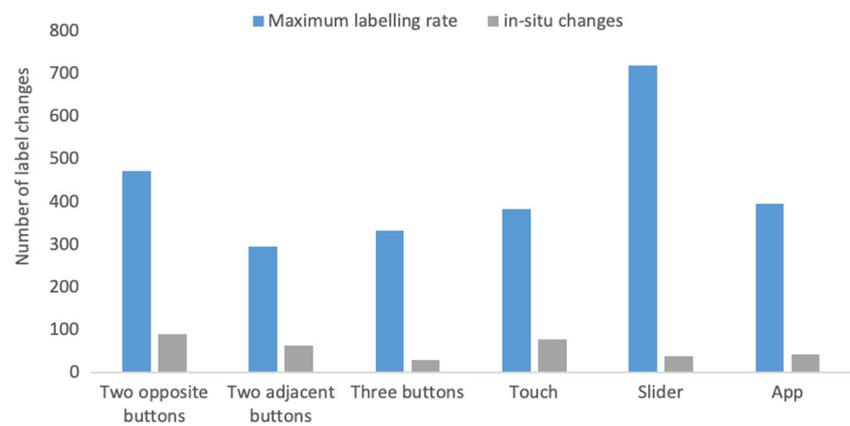

Fig. 4 Comparison of total maximum label changes per device

Figure 5 shows the total number of in situ recorded labels from all participants for each of the devices. Touch and slider have the highest total number of labels recorded as when using these labelling techniques, each label must be cycled through to change the label. Two opposite buttons had the smallest number of labels which is to be expected as a delay had to be added after a button is pressed to prevent incorrect labelling. Because of the delay, it was expected that the two adjacent buttons would similarly have a low data rate, but it achieved a higher rate than three buttons, possibly, because of the difficulty of accessing the three different buttons on different faces of the cube. This shows the position of the buttons has a greater impact on the number of labels recorded than the number of labelling interfaces embedded into the device. The comparative mobile app performed better than the buttoned devices but not as well as the slider or touch interfaces demonstrating the benefit of TUIs when a high labelling rate is required.

While all interfaces recorded more *walking* labels than any other label as expected due to the set route having more walking than stairs, the app had the fewest downstairs labels recorded demonstrating the difficulty in accessing virtual buttons in the same way as physical buttons where the button's position can have a major impact on its ease of use. Similarly, two adjacent buttons had a smaller proportion of upstairs and downstairs labels which is surprising as these labels are the easiest to access (by clicking a single button) compared with labelling walking that required both buttons to be pressed simultaneously. It is also likely that touch and slider have more downstairs labels than upstairs labels as downstairs must first be cycled through to reach either the walking or upstairs label.

## 5 Algorithms

In order to identify the three activities from the sensor data collected, deep neural networks were used to develop three predictive models. The performance of the three supervised, deep learning algorithms were tested to classify the sensor data into three activity classes. A multilayer recurrent neural network (RNN) [34] with long short-term memory (LSTM) [35], a multilayer RNN with gated recurrent unit (GRU) [36] and multilayer RNN with a stacked LSTM-GRU were selected due to their high performance and capabilities in classifying time-series data.

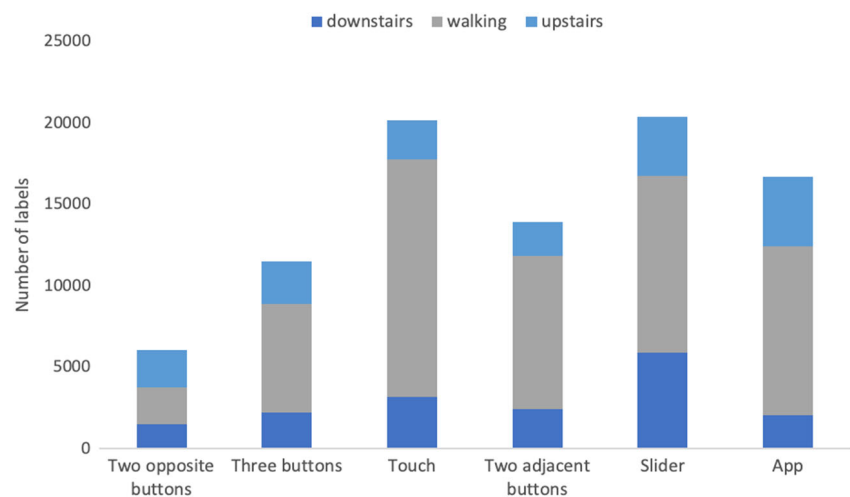

Fig. 5 Total number of recorded in situ labels for each device

It is vital to utilise LSTM or GRU cells when working with sequential data such as human activity and RNNs to capture long-term dependencies and remove the vanishing gradient. Recently the use of GRU cells is becoming increasingly popular due to its simpler design, which uses only two gates, a reset gate and an update gate rather than the three gates used by an LSTM: a forget gate, an input gate and an output gate. The use of a GRU cells can significantly reduce the time required to train models because of its simpler structure exposing the full hidden content to the next cell. GRU models have also been shown to outperform LSTM networks when there is a smaller training dataset, but LSTM models remember longer sequences than GRU models outperforming them in tasks requiring modelling long-distance relations [36–39]. Figure 6 shows the differences between the LSTM and GRU cells. Meanwhile, the stacked model will explore whether combining the LSTM and GRU cells within a single network improves or degrades performance in comparison with the base models. Batch normalisation was used on all models to normalise the inputs of each layer, so they have a mean of 0 and standard deviation of 1; this enables the models to train quicker, allows for higher learning rates and makes the weights easier to initialise [40].

The dataset collected from each of the five interfaces and mobile app was used to train the three models over 10 epochs with 10-fold cross-validation. The initial learning rate of the model was set to 0.0025 and a batch size of 32. The data sequences used during training have a length of $T = 100$ with an overlap of 20. Figure 7 shows the accuracy of each model. The stacked LSTM-GRU displayed little impact compared with the LSTM. Meanwhile, the GRU outperformed the LSTM and stacked models for most labelling techniques with the exception of two adjacent buttons where the LSTM network achieved the highest accuracy of all the labelling techniques at 92.8%. The overall GRU accuracies ranged between 68.5 and 89% demonstrating the impact different labelling techniques have on a dataset and thus the accuracy of a classification model.

The two adjacent buttons labelling technique achieved the highest accuracy of all the devices which is unexpected due to its complex nature where 2 buttons represent 3 labels. The second most accurate device, touch, was also unexpected due to the more complex interaction required of pressing the device using varying levels of force to record the different labels. It is possible that the more complex action forced users to have a greater focus on labelling their activity, resulting in more accurate labelling. This however may not be sustained if the device was to be used for several days. Even though three buttons and the slider labelling techniques resulted in the lowest changing labelling rate, they achieve consistently high accuracies in the three trained models. This demonstrates that although it may be more difficult to collect fast-changing data with these techniques, the collected data is reliable and capable of producing accurate classification models. The mobile app again performed moderately achieving 77.8% accuracy which although is not as high as touch, two adjacent buttons or three buttons; it is greater than slider and two opposite buttons.

Figure 8 shows the accuracy and loss of the combined user test data for all of the labelling interfaces during each epoch when trained using the RNN with GRU. The loss for each of the models gradually decreases, but the loss for the touch and slider decrease significantly as would be expected due to these interfaces achieving the highest overall accuracy.

It is possible that the datasets may contain potential biases, for example, if one user was particularly poor as labelling with one device; it may significantly impact the quality of the training dataset. To evaluate potential bias, the GRU model was trained using the data from five users using each interface as shown in Fig. 9.

There are extremely wide variations in the model accuracy ranging from 33.3 to 97.1%. Two opposite buttons and three buttons demonstrate the widest variation in model accuracy with accuracies reduced to 42.9% for user 1 using two opposite buttons and 33.3% for user 1 using three buttons. As the lowest accuracies were all performed by the same user, it indicates that this user experienced more difficulty using the interfaces than the other users. However, two opposite buttons also demonstrated poor accuracy (42.9%) when trialled by user 5; thus, it is shown that this interface results in poorer data collection as the data from the same user achieved consistently high accuracies for all other interfaces ranging between 74.2 and 87.5%. When comparing average model accuracy for each user, it shows some users can result in significantly better data collection and therefore model accuracy; for example, the overall accuracy between all interfaces for user 2 was 85.4%. The mobile app, two adjacent buttons, touch and slider all achieved high levels of accuracy when tested with each user demonstrating the reliability for those interfaces to consistently collect accurately labelled data. The touch interface achieved the highest overall accuracy at 97.1% when trained using data collected by user 4 although the data from the other interfaces collected by user 4 did not result in as higher accuracy demonstrating that user preference and familiarity with an interface plays an important role in the quality of data collected.

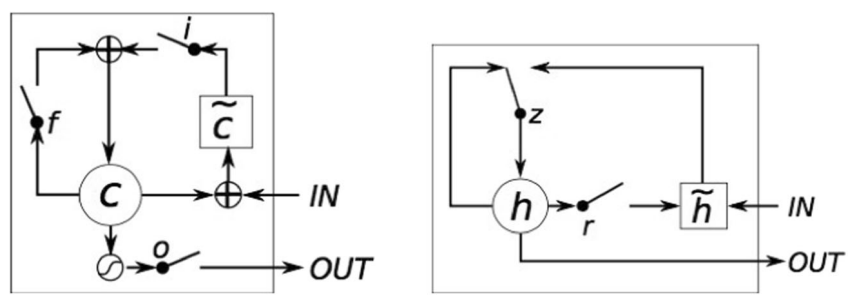

Fig. 6 Comparison of LSTM (left) and GRU (right) cells

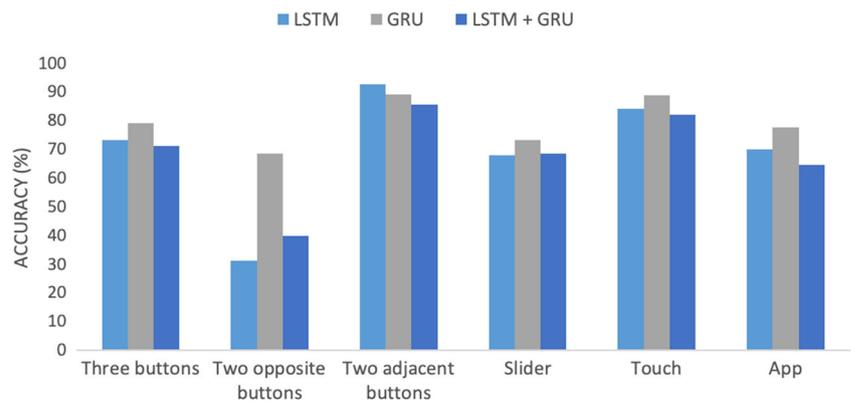

**Fig. 7** Comparison of deep learning techniques on the combined data collected from each devices

Classification accuracy alone does not provide an informed overview of the most beneficial labelling technique. The f1 score, a harmonic average of the precision and recall, for each label and device has been calculated, as shown in Table 2. Overall, the walking label has consistently higher precision and recall compared with the upstairs label which has the lowest f1 scores. The mobile app demonstrates good precision and recall when classifying upstairs but extremely poor precision and recall when classifying downstairs, potentially due to more mislabelling occurring when labelling climbing downstairs. The slider, two adjacent buttons and touch show the highest f1 scores which demonstrate their consistency as a useful labelling technique. Even though three buttons had a higher accuracy than slider, its f1 score is extremely low when labelling "upstairs", demonstrating its unreliability in classifying this class.

Cochran's Q test was performed to evaluate the three different models ($L = 3$) for each labelling technique providing a chi-square value and Bonferroni-adjusted $p$ value as shown in Table 3. Cochran's Q test is used to test the hypothesis that there is no difference between the classification accuracies across multiple classifiers distributed as chi-square with L-1 degrees of freedom. Cochran's Q test is similar to one-way repeated measures ANOVA and Friedman's test but for dichotomous data as the classification will either be correct or incorrect and can be applied across more than two groups unlike McNemar's test [41].

Assuming a significance level of = 0.05, Cochran's Q test shows for touch, two adjacent button, three buttons and the mobile app; the null hypothesis can be rejected as all three classifiers perform equally well. For the remaining labelling techniques, the null hypothesis has failed to be rejected showing there is a significant difference for the classifiers on those datasets. The $F$ test was also performed to compare the three classifiers as it is regarded analogous to Cochran's Q test. Assuming the same level of significance, the slider rejects the null hypothesis in addition to two adjacent buttons confirming Cochran's results.

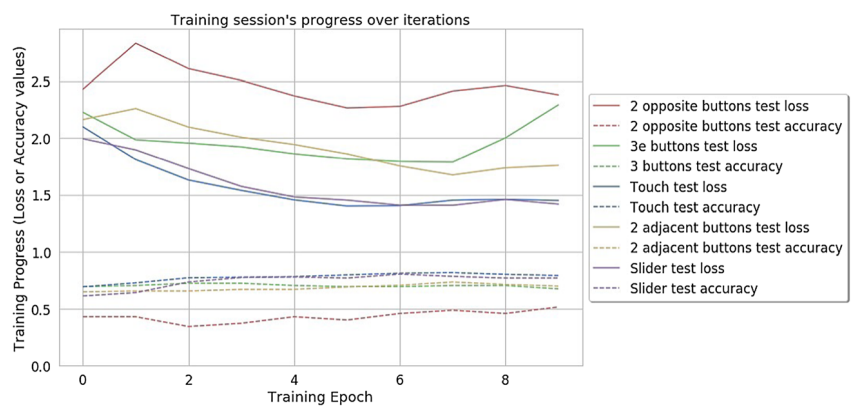

**Fig. 8** Comparison of training accuracy and loss when using GRU on the total data collected for each device

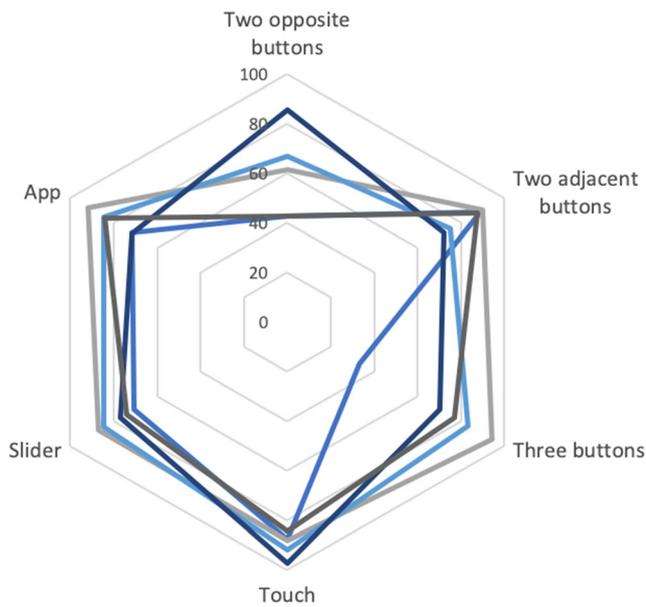

Fig. 9 Model accuracy when individually trained on 5 users' data

Table 3  Cochran's test and $F$ test comparing classification models

|  | Cochran's Q Chi$^2$ | Cochran's Q $p$ value | $F$ test | $F$ test $p$ value |
|---|---|---|---|---|
| Slider | 1.4 | 0.498 | 0.699 | 0.498 |
| Two adjacent buttons | 7.167 | 0.028 | 3.76 | 0.026 |
| Touch | 7.457 | 0.025 | 3.729 | 0.025 |
| Three buttons | 6.143 | 0.046 | 3.136 | 0.046 |
| Two opposite buttons | 2.533 | 0.282 | 1.277 | 0.285 |
| App | 13.241 | 0.001 | 6.852 | 0.001 |

Cochran's Q test shows there is a significant difference between the three models when trained on the two opposite buttons and slider datasets but does not show where the differences lie. To see which models contain the significant differences, the McNemar's test was performed to compare the predictive accuracy of each model using the 2 datasets.

Table 4 shows the resulting $p$ values when McNemar's test was performed. There is a significant difference between all of the models for both two opposite buttons and slider with the largest difference being between LSTM and the stacked network for both datasets. This demonstrates that both the labelling technique and the network architecture result in significant differences in the models' accuracy and reliability.

## 6 Discussion

To ensure the effectiveness of the labelling techniques, it is also vital to gain users' preference. Fifty users were asked which labelling technique they preferred. Figure 10 shows the results from the 50 users with 22% preferring the three buttons as it was simple to understand and use due to their being one label per button although this labelling technique did not result in accurate models. Similarly, 22% of people preferred two adjacent buttons with the mobile app following which is surprising as majority of the people are familiar with mobile apps, so it would be expected to be the most popular. The users found three buttons and two adjacent buttons to be simpler to operate than the mobile app due to the physical buttons being quicker and easier to press than the virtual button on the app, which were often missed. Two opposite buttons followed again possibly due to the simplicity and familiarity of the buttons to label data. The slider was well received, but the granular control made the middle label more difficult to access meaning careful consideration had to be made to ensure actions were being correctly labelled. Finally, the fewest number of people preferred the touch-based labelling technique due to the complexity of having to touch with varying levels of pressure to correctly label the data. However, touch did result in highly accurate models showing that while the increased attention required is not preferred, it does ensure accurate data labelling, but this may not be sustained over the long periods.

Table 2  F1 score for each label when trained using each device

|  | Downstairs | Walking | Upstairs |
|---|---|---|---|
| Slider | 0.7 | 0.82 | 0.69 s |
| Two adjacent buttons | 0.82 | 0.91 | 0.75 |
| Touch | 0.69 | 0.94 | 0.83 |
| Three buttons | 0.59 | 0.8 | 0.3 |
| Two opposite buttons | 0.58 | 0.75 | 0.42 |
| App | 0.23 | 0.60 | 0.82 |

Table 4  McNemar's test comparing 2 opposite buttons and slider

|  | Two opposite buttons | | | Slider | | |
|---|---|---|---|---|---|---|
|  | GRU | LSTM | Stacked | GRU | LSTM | Stacked |
| GRU | BA | 0.228 | 0.125 | NA | 0.286 | 0.596 |
| LSTM | 0.228 | NA | 0.546 | 0.286 | NA | 0.845 |
| Stacked | 0.125 | 0.546 | NA | 0.596 | 0.845 | NA |

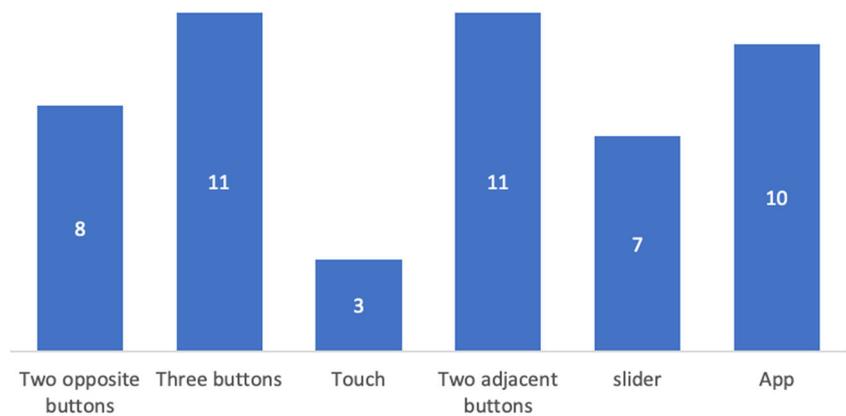

**Fig. 10** Comparison of 50 users' labelling preference

While the user preference of a given labelling technique does not correlate with the accuracy achieved for each method, it shows the benefits of using buttons as they are well-received by users and also achieve high accuracy. A lower number of buttons than labels was jointly preferred by the users and achieved the highest accuracy, but the number of buttons must remain equal to the number of labels to ensure the users do not experience confusion when labelling. The position of the buttons has also shown to impact on user preference. In terms of labelling rate and model accuracy, two adjacent buttons were preferred by users and resulted in 24.3% higher model accuracy than two opposite buttons which had a higher total number of recorded in situ labels but a lower labelling rate. It is imperative to balance user preference with the rate at which the data needs to be labelled and the accuracy is required from the model when selecting an appropriate labelling technique.

Novel labelling methods including the slider and touch displayed their own strengths and weaknesses. Labelling using touch resulted in high model accuracy and labelling rate but was the least favoured by users. If accurate labelling is required for only short periods, labelling via touch could be ideal. The slider was liked by the users and had the highest labelling rate but achieved the second worse accuracy of all the devices at 73.4% showing the slider is best for continually changing or granular data that would be more difficult to label with buttons.

Surprisingly the mobile app was not the most popular labelling technique even though all participants were more familiar with apps than the other interfaces. The data collected from the mobile app shows it achieved a moderate labelling rate and model accuracy despite participants' familiarity. A possible reason why the mobile app did not result in the most accurate data is that virtual buttons can be easier to miss than physical interfaces. However, when used in real world environments, apps are easier to deploy, but solely using an app does not allow for any additional sensors that are not embedded within the smartphone to be used. Apps possess many benefits when used to label motion data including ease of access, but when additional sensors are required, using apps for purely labelling is not recommended over physical labelling techniques.

One of the most significant challenges encountered was the inconsistent quality of labelled data, as when collecting in situ data to train machine learning models, it is not possible to ensure all users are successfully labelling their actions. For example, the wide variation in labelling rates was most likely due to users not following the set route as they were unaccompanied during the labelling process to better replicate in-situ data collection.

Additionally, as users had to repeat the experiment five times to enable them to use each device, their labelling rate may change as they become more familiar with the experiment. To combat this, users were provided with the devices in varying orders preventing the same device from being used by all users at the same stage of the experiment.

Overall, when labelled in situ sensor data is required, the use of physical labelling interfaces should be considered as they have demonstrated their ability to improve labelling rate, accuracy and user preference in comparison with mobile apps, which are most commonly used to label sensor data.

## 7 Applications and future work

AI-powered edge computing has numerous potential applications as it is not always possible to label real-time data using a smartphone application. Common uses for tangible labelling techniques include times when users may be engaged in other activities such as labelling while physically active. Additionally, tangible labelling techniques are required in cases where specialist sensors are required to collect labelled data such as physiological sensors used to label mental well-being or environmental sensors to measure pollution. The labelling techniques discussed provide new opportunities to label real-time sensor data that has traditionally been challenging to label. This data can then be used to train models, possibly on the device using edge computing to classify sensor data in real time.

In the future, these labelling techniques could be evaluated on other data types including the use of more specialist sensors to further prove their effectiveness. Additionally, longer data collection trials could be conducted as while this experiment was short, it demonstrates the requirements of using tangible labelling techniques to improve labelling rate and overall model accuracy.

## 8 Conclusion

Tangible user interfaces are ideal interfaces for data collection and running real-time machine learning classifiers, but first real-world-labelled data must be collected. Images, video and audio data can all be labelled offline, but this is not possible with time-series sensor data. To address this issue and collect in situ labelled sensor data, five different labelling techniques have been embedded into TUIs including two opposite buttons, two adjacent buttons, three buttons, slider, touch and a comparative mobile application. The interfaces were used by the participants to label three physical activities enabling the performance of each technique to be evaluated. It is vital to compare the different labelling techniques as machine learning models can only be as accurate as the labelled data they are trained on.During this pilot study, participants used six labelling interfaces to collect data that was used to train various RNNs. The results demonstrate that while a touch interface results in a high labelling rate and high model accuracy, it is the least favoured by the users due to the high level of attention required to use the device. The mobile app was popular with users due to its familiarity but only achieved the fourth highest accuracy. The slider resulted in high user preference and labelling rate but poor model accuracy, while two adjacent buttons achieved both high user preference and the highest model accuracy showing it is the most beneficial technique for this data collection.

Overall, this exploratory work demonstrates embedding labelling techniques within TUIs addresses many of the challenges facing the collection of in situ, time-series sensor data. When collecting labelled data, the nature of the data, labelling rate, duration of data collection and user preference all need to be considered to ensure the most effective labelling technique is used. This will increase the reliability of in situ labelled datasets and enable the development of more accurate machine learning classifiers.

**Funding information** Dr. Chamberlain's part in this work was supported by the Engineering and Physical Sciences Research Council (grant no. EP/T51729X/1) project RCUK Catapult Researchers in Residence award: Digital - Disruptive Beats - Music - AI - Creativity - Composition and Performance.